\documentclass[11pt]{article}
\usepackage[preprint]{acl}
\usepackage{times}
\usepackage{latexsym}
\usepackage[T1]{fontenc}
\usepackage[utf8]{inputenc}

\usepackage{amsmath,amsfonts,bm}

\def\eqref#1{equation~\ref{#1}}

\def\1{\bm{1}}

\DeclareMathAlphabet{\mathsfit}{\encodingdefault}{\sfdefault}{m}{sl}
\SetMathAlphabet{\mathsfit}{bold}{\encodingdefault}{\sfdefault}{bx}{n}

\usepackage{hyperref}
\usepackage{url}

\usepackage{amsmath}
\usepackage{amssymb}
\usepackage{booktabs}
\usepackage{array}
\usepackage{tabularx}
\usepackage{graphicx}
\usepackage{microtype}
\usepackage{placeins}
\usepackage{pifont}
\graphicspath{{figures/}}
\usepackage{multirow}
\usepackage[table]{xcolor}

\title{Ensemble-Aware Structural Encoding (EASE): Conditioning Protein Encoders on Peptide Conformational Ensembles}

\author{
Ruoyao Zhang$^{1}$ \quad
Hongxuan Li$^{2}$ \quad
Jie Peng$^{2}$ \quad
Tianlong Chen$^{2}$ \\
$^{1}$University of Waterloo \qquad
$^{2}$University of North Carolina at Chapel Hill
}

\begin{document}

\maketitle

\begin{abstract}
Protein foundation models provide transferable representations for downstream peptide prediction tasks. 
However, existing representations typically reduce a peptide to a single sequence or structure, which may miss conformational variability relevant to key peptide properties, including antibacterial activity, hemolysis, toxicity, stability, and binding.
For example, protease resistance correlates more strongly with conformational dynamics measured by hydrogen exchange than with average helicity.
Therefore, we introduce Ensemble-Aware Structural Encoding (EASE), a framework for effectively extracting structural information from conformational ensembles for peptide representation.
To extract useful information from noisy conformers, EASE selects central conformers representing a peptide's major structural states and fuses their individual representations with a permutation-invariant operator to produce a peptide-level representation for downstream prediction.
We then evaluate our framework on six peptide property-prediction tasks under leakage-controlled splits.
Against matched single-structure baselines, EASE improves performance consistently on six tasks with a mean absolute gain of about $3.0\%$ across two structural encoders. Gains keep growing as we add conformers through K=20. A single peptide conformer therefore provides a limited basis for structural representation, whereas our framework distilling and fusing multiple conformers yields a more informative representation.

\end{abstract}

\section{Introduction}
\label{sec:intro}

Peptides are short amino-acid chains, conventionally under fifty residues, that sit between small molecules and proteins in both size and pharmacology.
Their ability to recognize biological targets, along with advances in peptide synthesis and chemical modification, made them increasingly important in therapeutics \citep{henninot2018peptide, muttenthaler2021trends}.
Accordingly, recent representation learning methods aim to predict peptide properties such as biological activity, toxicity, and stability \citep{zhang2025pepland,guntuboina2023peptidebert,zhang2026pepbenchmark}.
Many of these methods build on pretrained Protein Language Models (PLMs), which have become a widely used approach for learning transferable representations from amino-acid sequences. Structure aware models, such as ESM-3, extend sequence representations with structural information \citep{hayes2025simulating}, but typically process a single structure at a time. This presents a unique challenge for peptides. Peptides are conformationally heterogeneous, and their bioactive conformations are often low-probability states that a single static structure cannot fully capture \citep{zhang2026alphafoldpeptide}. 
For example, protease resistance correlates more strongly with conformational dynamics measured by hydrogen exchange than with average helicity \citep{shi2013hydrogen}.

A conformational ensemble provides a more complete description of peptides' variability, but obtaining such ensembles traditionally required expensive molecular dynamics simulations \citep{karplus2002md,hollingsworth2018md}.
Recent generative models such as BioEmu, AlphaFlow, and Str2Str are trained on molecular dynamics to generate equilibrium-like structural ensembles directly, making the process affordable and scalable \citep{lewis2025bioemu,jing2024alphaflow,lu2024str2str}.
At the same time, recent studies suggest that conformational information can improve protein representations and downstream prediction \citep{shi2026ensembits, kalifa2026dynamicsplm}. 
However, existing peptide models primarily focus on sequence PLMs \citep{guntuboina2023peptidebert,chen2025pepmlm}, chemical representations \citep{feller2025peptideclm, adamczyk2026fingerprints}, or a single predicted structure \citep{zhao2024deepampnet,aguilerapuga2024structure,liu2026pepanno}, leaving the value of conformational ensembles for peptide representation learning largely unexplored. 
Importantly, simply generating more structures does not directly yield a better representation, as conformational ensembles can contain substantial redundancy, motivating the extraction of a compact subset of conformers that captures their structural diversity \citep{conev2023engens}.
This presents two related questions: how to effectively extract useful information from noisy conformational data, and to what extent such information can complement representations learned by PLMs.

To address these questions, we introduce Ensemble-Aware Structural Encoding (EASE), a general framework for extracting predictive structural information from conformational ensembles. From a large sampled conformers set, EASE performs structural clustering and selects a subset containing central conformers. Each representative conformer passes through a shared and adapted structural encoder, and their representations are fused through a permutation-invariant operator to obtain a single peptide representation.
Our contributions include: 
\ding{182} \textbf{Ensemble-Aware Structural Encoding (EASE).} We introduce a framework that systematically exploits conformational diversity of peptides, producing representations that improve downstream prediction.
\ding{183} \textbf{Consistent gains across structural encoders and peptide properties.} Across two distinct pretrained structural encoders, GearNet-Edge and ESM-3, and six downstream tasks, EASE outperforms a matched single-structure baseline by $+3.8\%$ on GearNet-Edge across all six tasks and $+2.2\%$ on ESM-3 across five of them.
\ding{184} \textbf{Analysis of how ensemble information should enter the model.} We systematically compare EASE against six alternative mechanisms for injecting ensemble-derived summary statistics into the model. EASE provides the most consistent improvement, while the alternatives recover only a limited portion of the gain.
\ding{185} \textbf{Analysis of ensemble information and protein language models.} We examine when conformational information provides additional predictive value beyond Protein Language Model (PLM) representations, through comparisons across PLM, structural, and PLM-conditioned structural encoders.


\section{Related work}
\label{sec:related}

\subsection{Peptide representation learning}
Existing peptide models are mainly based on three approaches.
Sequence-based approaches adapt protein language models to the peptide domain or train peptide-specific models on sequences, chemical representations use notations such as SMILES and HELM to non-standard amino-acids, and structure-aware approaches further incorporate geometric information from 3D structures \citep{fernandezdiaz2024autopeptideml,guntuboina2023peptidebert,feller2025peptideclm,lee2025helmbert,zhang2025pepland,zhao2024deepampnet,liu2026pepanno,yu2026samtp}. 
As a sequence-based example, PeptideBERT fine-tunes ProtBERT directly on downstream peptide properties prediction such as hemolysis and solubility \citep{guntuboina2023peptidebert}.
For chemical representations, PeptideCLM learns from SMILES strings, enabling it to represent non-canonical residues and chemical modifications beyond standard amino-acid vocabularies \citep{feller2025peptideclm}.
For structure-aware learning, deepAMPNet combines sequence embeddings with AlphaFold2-predicted structures through a graph neural network to identify antimicrobial peptides \citep{zhao2024deepampnet}.
Despite their different input modalities, most pretrained models consume a single sequence, chemical string, or structural conformation at a time.
In contrast, our approach takes into account multiple conformers of the same peptide to learn an ensemble-informed representation.

\subsection{Learning from conformational ensembles}
Representation learning over conformational ensembles has been explored primarily on small molecules and later extended to peptides and proteins \citep{axelrod2023conformer,zhu2024marcel,nguyen2026facet,feller2026ensembleegnn,wu2022protmd,rollins2024ablef,kalifa2026dynamicsplm,shi2026ensembits}.
Axelrod and G\'omez-Bombarelli encode multiple conformers for small molecules and apply attention over their representations \citep{axelrod2023conformer}. Additionally, MARCEL systematically evaluates conformer ensembles across molecular GNNs using aggregation methods such as mean pooling, DeepSets, and self-attention \citep{zhu2024marcel}.
The work most closely related to our peptide setting is EnsembleEGNN, which encodes cyclic peptide conformers with shared EGNN layers, applies attention over them, and pretrains the resulting ensemble model on cyclic peptide conformations before downstream permeability task \citep{feller2026ensembleegnn}.
As an example of protein representation models using ensembles, DynamicsPLM combines structural tokens from multiple conformers at each residue before processing them with a protein language model \citep{kalifa2026dynamicsplm}.
Alternatively, ENSEMBITS learns discrete tokens that represent the conformational ensemble per residue \citep{shi2026ensembits}.
In contrast to these works, EASE additionally considers how a large sampled ensemble should be distilled before representation learning.
Additionally, it addresses whether existing pretrained single-structure encoders can be adapted to learn peptide representations from conformational ensembles without ensemble-specific pretraining. 
Our inspiration came from multiple-instance learning, in which representations are learned from bags of instances through permutation-invariant aggregation \citep{zaheer2017deep,ilse2018attention}.
We adopt a similar idea, providing an encoder-agnostic way to extract ensemble-informed peptide representations for downstream prediction.

\section{Method}
\label{sec:method}

\subsection{Pipeline Overview}
\label{sec:mce}

\begin{figure*}[t]
\centering
\includegraphics[width=\textwidth]{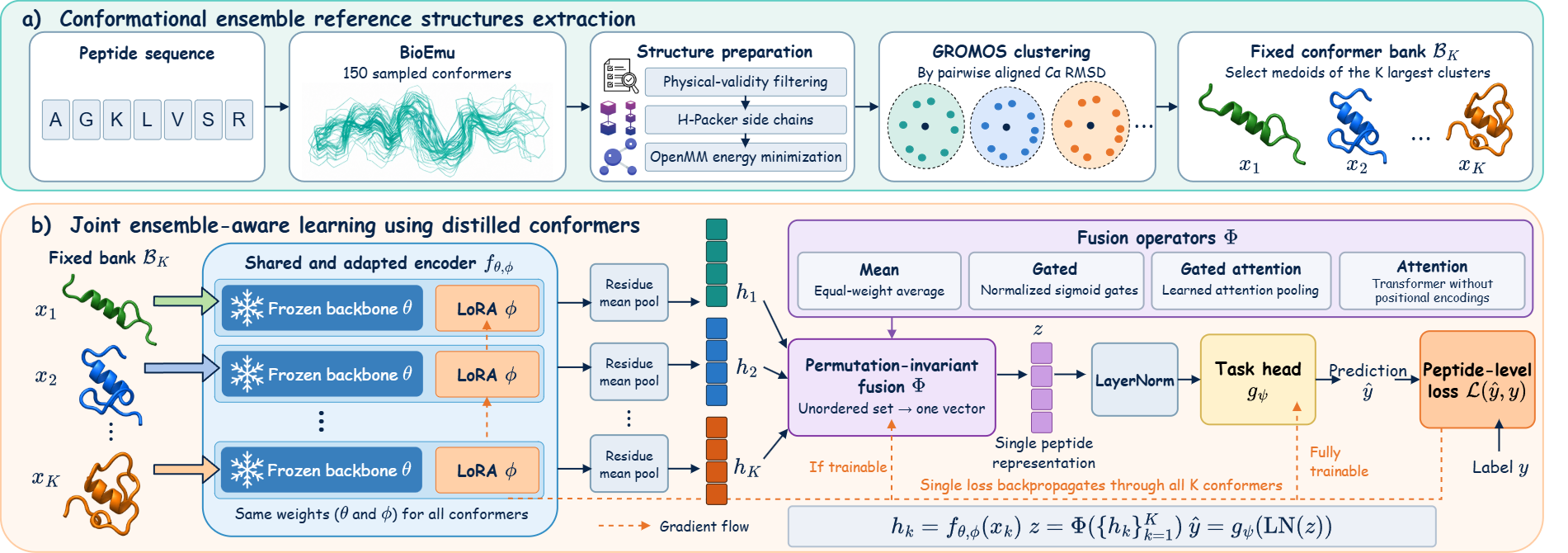}
\caption{\small Ensemble-Aware Structural Encoding pipeline.}
\label{fig:pipeline}
\end{figure*}

Let a peptide's conformational ensemble $\mathcal{E}$ be an unordered, unweighted, and variable-size set of conformer structures, generated by an ensemble sampler.
We perform structural clustering to select a fixed bank $\mathcal{B}_K \subset \mathcal{E}$ of $K$ conformers from $\mathcal{E}$. 
Each conformer $x_k$ passes through a shared adapted encoder trunk $f_{\theta,\phi}$, where pretrained weights $\theta$ are frozen and low-rank adapters $\phi$ are shared across conformers. Their representations are then fused to produce a single loss per peptide:
\begin{equation*}
\begin{aligned}
h_k &= f_{\theta,\phi}(x_k), \quad k = 1, \dots, K; \\
z &= \Phi(h_1, \dots, h_K); \\
\hat{y} &= g_\psi(\mathrm{LN}(z)),
\end{aligned}
\end{equation*}
where $\Phi$ is a permutation-invariant fusion operator (\S\ref{sec:fusion}) and $g_\psi$ is the task head.
Because the adapters are shared, gradients from all $K$ conformers contribute to each update of $\phi$.

To test whether EASE transfers across structural representations, we evaluate it with two pretrained structural encoders:
\ding{182} \textbf{GearNet-Edge.} Each conformer is represented as a relational residue graph. Per-residue outputs are mean-pooled over valid positions to obtain $h_k \in \mathbb{R}^{512}$. LoRA adapters \citep{hu2022lora} are applied to the linear and self-loop maps of every node and edge layer, with one factor zero-initialized so the adapted model matches the pretrained encoder at initialization.
\ding{183} \textbf{ESM-3.} Each conformer is tokenized by ESM-3's structure track and combined with the peptide's sequence track. The sequence track outputs remain the same across the $K$ inputs so only the structure track varies. Per-residue outputs over valid positions are mean-pooled to obtain $h_k \in \mathbb{R}^{1536}$, which is then normalized before the fusion operator.

\subsection{Ensemble generation}
As the first step of the pipeline, we generate peptide conformational ensembles $\mathcal{E}$ using BioEmu \citep{lewis2025bioemu}, sampling 150 conformers per peptide.
After applying BioEmu's physical-validity filtering, we retain a median of 145 conformers (range: 72--150; all $\geq 50$) per peptide.
Since BioEmu generates backbone-frame structures only, we reconstruct side chains with H-Packer \citep{mcpartlon2023hpacker} and perform local energy minimization in OpenMM \citep{eastman2017openmm} using Amber99SB \citep{hornak2006amber99sb} and OBC2 implicit solvent \citep{onufriev2004obc}.

\subsection{Conformer banks and single-structure baseline selection method}
\label{sec:banks}

The next step is selecting the $K$ conformers from $\mathcal{E}$. For each peptide, we compute pairwise C$\alpha$ RMSD between all sampled conformers after pairwise Kabsch alignment. We apply GROMOS clustering \citep{daura1999peptide}, where the conformer with the most neighbors within a cutoff forms a cluster with those neighbors; the selected cluster is then removed and the procedure repeats. We then take the medoid (the conformer with the lowest mean RMSD to all other conformers in the cluster) of each of the $K$ largest clusters. 
To accommodate differences in peptide length and flexibility, the GROMOS clustering cutoff is defined relative to each peptide's pairwise-RMSD distribution. We progressively use quantiles $\{0.25,0.15,0.10,0.05,0.02,0.01\}$ until at least $K$ clusters are obtained.
\footnote{Across 900 peptides sampled from six tasks, 81\% yield at least $K$ clusters at the first cutoff and 100\% at the second. }

\subsection{Fusion operators}
\label{sec:fusion}

After each conformer in $\mathcal{B}_K$ passes through the encoder, the output embeddings enter a fusion operator. We compare four permutation-invariant fusion operators $\Phi$, mapping $K$ $d$-dimensional embeddings to a single $d$-vector so that the downstream head is identically shaped across arms. 
\ding{182} \textbf{Mean.}
Parameter-free mean pooling weights all conformers equally:
$\Phi(h_1,\dots,h_K)=\frac{1}{K}\sum_k h_k$.
\ding{183} \textbf{Gated.}
Each conformer receives a scalar gate, adding $d+1$ parameters:
$s_k=\sigma(u^\top h_k+b)$,
$w_k=s_k/\sum_j s_j$, 
$\Phi=\sum_k w_k h_k$.
\ding{184} \textbf{Gated attention.}
We use the multiple-instance-learning pooling of \citet{ilse2018attention}:
$s_k=w^\top[\tanh(Vh_k)\odot\sigma(Uh_k)]$,
$a_k=\exp(s_k)/\sum_j\exp(s_j)$,
$\Phi=\sum_k a_k h_k$,
where $V,U\in\mathbb{R}^{L_a\times d}$. We use the $L_a=128$ from \citet{ilse2018attention}.
\ding{185} \textbf{Attention.}
We apply one Transformer encoder layer over the conformer axis without positional encodings and mean-pool its outputs. The layer is therefore permutation equivariant, and mean pooling makes $\Phi$ permutation invariant. Due to large added parameters, we reduce the width $d$ before the Transformer layer, down-projecting to 64 for GearNet-Edge and 128 for ESM-3.
Table ~\ref{tab:fusion-params} in the Appendix presents a summary of added parameters by each fusion operator.
\section{Experimental setup}
\label{sec:setup}

\noindent\textbf{Tasks.}
We evaluate on six peptide property-prediction tasks (Table~\ref{tab:tasks}), comprising five peptide-intrinsic tasks and one peptide-protein interaction task.
The tasks span key endpoints in peptide development, including safety (toxicity and hemolysis), efficacy (E. coli minimum inhibitory concentration), pharmacokinetics (blood stability), and target engagement (peptide-protein interaction).
Five tasks are drawn from PepBenchmark \citep{zhang2026pepbenchmark}; blood stability is drawn from PepMSND \citep{hu2025pepmsnd}.
For all tasks, we restrict peptides to canonical amino acids with lengths up to 50 residues.
Two tasks include auxiliary inputs in addition to the peptide representation.
For peptide-protein interaction, we represent the receptor using a frozen ESM-2-150M encoder, mean-pooled and fixed at 640 dimension \citep{lin2023esm2}. Receptor representation is concatenated with the peptide representation before the prediction head.
For blood stability, assay conditions, including species and in vivo vs. in vitro, are encoded as a six-dimensional one-hot vector and concatenated before the prediction head.
These auxiliary inputs are held fixed across all experiments, ensuring that only the peptide representation changes for each ablation.

\smallskip
\noindent\textbf{Splits.}
The five single peptide tasks use a homology-controlled five-fold cross validation.
Sequences are clustered with MMseqs2 \citep{steinegger2017mmseqs2} at 50\% identity, and entire sequence-similarity components are assigned to folds to minimize homology leakage.
In each run, three folds are used for training, one for validation, and one for testing, with each fold serving as the test set once.
For peptide--protein interaction, we use PepBenchmark's released peptide-cold splits, which ensure that test peptides do not occur in the corresponding training set.



\begin{table}[t]
\centering
\caption{\small The six peptide property-prediction tasks.}
\label{tab:tasks}
\small
\setlength{\tabcolsep}{4pt}
\begin{tabular}{@{}lccl@{}}
\toprule
\textbf{Task} & \textbf{Peptides} & \textbf{Metric} & \textbf{Application} \\
\midrule
Toxicity
    & 4,056 & AUROC & Safety \\
Hemolytic
    & 4,259 & AUROC & Blood compatibility \\
Hemolytic HC50
    & 1,926 & RMSE & Blood compatibility \\
\emph{E. coli} MIC
    & 3,161 & RMSE & AMP activity \\
Blood stability
    & 406 & AUROC & Proteolytic stability \\
PepPI
    & 5,931 & AUROC & Target binding \\
\bottomrule
\end{tabular}
\end{table}

\smallskip
\noindent\textbf{Baselines.}
\label{sec:baselines}
To evaluate our method, we compare EASE with several baselines.
\ding{182} \textbf{Reference conformer.} We define the reference conformer as the medoid conformer from each peptide's $K=10$ bank, as stated in \S\ref{sec:banks}. This baseline uses the same encoder, adapters, prediction head, and training protocol as the EASE method, but receives only the reference conformer instead of the full $K$ conformer bank. 
\ding{183} \textbf{PLM baseline.} We consider a sequence-only baseline as a conventional reference against protein language model representations for each encoder. For ESM-3, this baseline masks the structure track but keeps the sequence track with all other model and training settings unchanged. Since GearNet-Edge is fully structural, we use ESM-2 \citep{lin2023esm2} with the same LoRA configuration and training protocol, selecting the 35M variant as the closest in size to the 20.2M parameter GearNet-Edge model.
\ding{184} \textbf{Chemical representation baseline.} To offer a comparison with SMILES-based chemical models, we additionally evaluate the downstream tasks on ChemBERTa-77M \citep{chithrananda2020chemberta}, a pretrained molecular language model.

\smallskip
\noindent\textbf{Ablation studies on the ensemble size.}
To study the effect of the ensemble size on downstream performance, we conduct ablation studies on $K$, the size of the selected conformer bank. We evaluate $K\in\{5,10,20\}$ while holding the reference conformer fixed. For each $K$, the bank consists of the reference conformer and $K-1$ cluster medoids selected by running the same GROMOS clustering procedure (\S\ref{sec:banks}) on the remaining conformer pool.

\smallskip
\noindent\textbf{Training protocol.}
Hyperparameters are selected on validation performance separately for the two encoders. A table of hyperparameters are provided in the Appendix.
For GearNet-Edge, we tune LoRA rank and prediction-head dropout using a grid shared across comparison arms and select one configuration per task, preventing arm-specific retuning.
For ESM-3, we tune the trunk learning rate and batch size separately for each task and comparison arm.
In all experiments, the checkpoint with the best validation performance is used for test evaluation.
GearNet-Edge experiments use five folds and four training seeds per fold, and ESM-3 experiments use five folds and three seeds per fold.
Seeds vary parameter initialization and minibatch ordering within each fixed split.

\smallskip
\noindent\textbf{Statistics reporting.}
We report test performance as mean $\pm$ standard deviation over fold averages, with training seeds averaged within each fold.
For comparisons with the baselines, we first average over seeds within each fold and compute paired differences between methods.
We report the mean paired difference and the number of folds in which the EASE method improves over the baselines.

\section{Results}
\label{sec:results}

\subsection{EASE improves over a single structure}
\label{sec:res-main}

Table~\ref{tab:sequence} reports absolute test performance for both encoders with all baselines specified in \S\ref{sec:baselines}. Table~\ref{tab:fusion-paired} reports paired differences between EASE with mean fusion over ten conformers and the matched reference conformer baseline defined in \S\ref{sec:baselines}.
Averaged across the six tasks, EASE improves over the baseline by $+0.0274$ with GearNet-Edge and $+0.0162$ with ESM-3.
Specifically, on GearNet-Edge, EASE improves over all six tasks and 28 of 30 folds.
On ESM-3, EASE improves on five of six tasks over 23 of 30 folds. The exception is blood stability, where EASE underperforms the reference conformer by $-0.0108$ on average and loses on all five folds. 
A plausible explanation is the task's much smaller dataset compared to the other tasks, making its fold-to-fold performance volatile for both encoders.

Excluding blood stability, the remaining five tasks show a similar average improvement for both encoders ($+0.0248$ for GearNet-Edge, $+0.0216$ for ESM-3). This similar improvement over most tasks despite the choice of encoders is notable because GearNet-Edge and ESM-3 represent structure differently.
GearNet-Edge directly builds a relational graph over the residues, whereas ESM-3 uses a VQ-VAE and represents local structure through discrete structure tokens.
The task-level gains are less consistent across encoders. For example, GearNet-Edge gains more on peptide-protein interaction ($+0.0267$ versus $+0.0191$) and HC50 regression ($+0.0321$ versus $+0.0238$), while ESM-3 gains more on E. coli MIC regression ($+0.0327$ versus $+0.0196$).
Thus, the two encoders agree on the overall benefit of multi-conformer encoding, but not on which tasks benefit most.

\begin{table*}[t]
\centering
\caption{\small Absolute test performance. Mean $\pm$ standard deviation over folds $\times$ training seeds. AUROC for classification and RMSE for regression. Average rank is obtained by ranking all arms on each task and averaging their ranks across the six tasks.}
\label{tab:sequence}
\resizebox{\textwidth}{!}{%
\begin{tabular}{ll|cccccc|c}
\toprule
\textbf{Encoder} & \textbf{Arm}
& \textbf{Toxicity} $\uparrow$
& \textbf{Hemolytic} $\uparrow$
& \textbf{HC50} $\downarrow$
& \textbf{\emph{E. Coli} MIC} $\downarrow$
& \textbf{Stability} $\uparrow$
& \textbf{Interaction} $\uparrow$
& \textbf{Avg. rank} $\downarrow$ \\
\midrule
\multirow{3}{*}{\textbf{ESM-3}}
& Sequence track only
& 0.7896 $\pm$ 0.0640
& 0.9010 $\pm$ 0.0245
& 0.5540 $\pm$ 0.0328
& 0.6667 $\pm$ 0.0294
& 0.6983 $\pm$ 0.0231
& 0.8308 $\pm$ 0.0121
& 3.33 \\
& Reference conformer
& 0.7792 $\pm$ 0.0534
& 0.9013 $\pm$ 0.0209
& 0.5793 $\pm$ 0.0230
& 0.6842 $\pm$ 0.0225
& \textbf{0.7141 $\pm$ 0.0266}
& 0.8329 $\pm$ 0.0118
& 3.08 \\
\rowcolor{gray!10}
& EASE, mean $K=10$
& \textbf{0.8058 $\pm$ 0.0387}
& \textbf{0.9071 $\pm$ 0.0161}
& 0.5554 $\pm$ 0.0334
& \textbf{0.6515 $\pm$ 0.0183}
& 0.7032 $\pm$ 0.0314
& \textbf{0.8520 $\pm$ 0.0100}
& \textbf{1.67} \\
\midrule
\multirow{3}{*}{\textbf{GearNet-Edge}}
& ESM-2 (35M)
& 0.7652 $\pm$ 0.0597
& 0.8936 $\pm$ 0.0184
& \textbf{0.5505 $\pm$ 0.0254}
& 0.6657 $\pm$ 0.0205
& 0.6558 $\pm$ 0.0694
& 0.8061 $\pm$ 0.0073
& 4.00 \\
& Reference conformer
& 0.6893 $\pm$ 0.1008
& 0.8617 $\pm$ 0.0221
& 0.6460 $\pm$ 0.0191
& 0.6847 $\pm$ 0.0202
& 0.6737 $\pm$ 0.0458
& 0.8059 $\pm$ 0.0159
& 6.33 \\
\rowcolor{gray!10}
& EASE, mean $K=10$
& 0.7231 $\pm$ 0.0938
& 0.8733 $\pm$ 0.0225
& 0.6139 $\pm$ 0.0164
& 0.6651 $\pm$ 0.0275
& \textbf{0.7141 $\pm$ 0.0525}
& 0.8327 $\pm$ 0.0101
& 3.92 \\
\midrule
& ChemBERTa & 0.6473 $\pm$ 0.0922
& 0.8677 $\pm$ 0.0247
& 0.6003 $\pm$ 0.0251
& 0.7094 $\pm$ 0.0251
& 0.5773 $\pm$ 0.1116
& 0.8351 $\pm$ 0.0156
& 5.67
\\
\bottomrule
\end{tabular}}
\end{table*}

\subsection{Simple mean fusion is sufficient for EASE}
\label{sec:res-fusion}
\label{res-fusion} 

In this section, four permutation-invariant fusion methods are compared as stated in \S\ref{sec:fusion}.
Table ~\ref{tab:fusion-paired} reports their paired differences over the same folds against the reference conformer baseline, and the absolute performance can be found in the Appendix ~\ref{tab:fusion-abs}.
Mean pooling performs as well as or better than the learned fusion operators overall.
On ESM-3, its average improvement over the baseline is $+0.0162$, compared with $+0.0140$ for the scalar gate, $+0.0111$ for transformer attention, and $+0.0108$ for gated attention.
On GearNet-Edge, the scalar gate is marginally higher on average ($+0.0285$ versus $+0.0274$), while gated attention and transformer attention reach $+0.0268$ and $+0.0186$, respectively.
Across both encoders, no learned operator improves over mean pooling by more than $0.0011$ on average.
We note that mean pooling introduces no additional parameters.
Therefore, the improvement of the mean fusion operator cannot be explained by additional fusion capacity, and the principal change is that the model receives multiple conformers instead of one.

\begin{figure}[t]
    \centering
    \includegraphics[width=\linewidth]{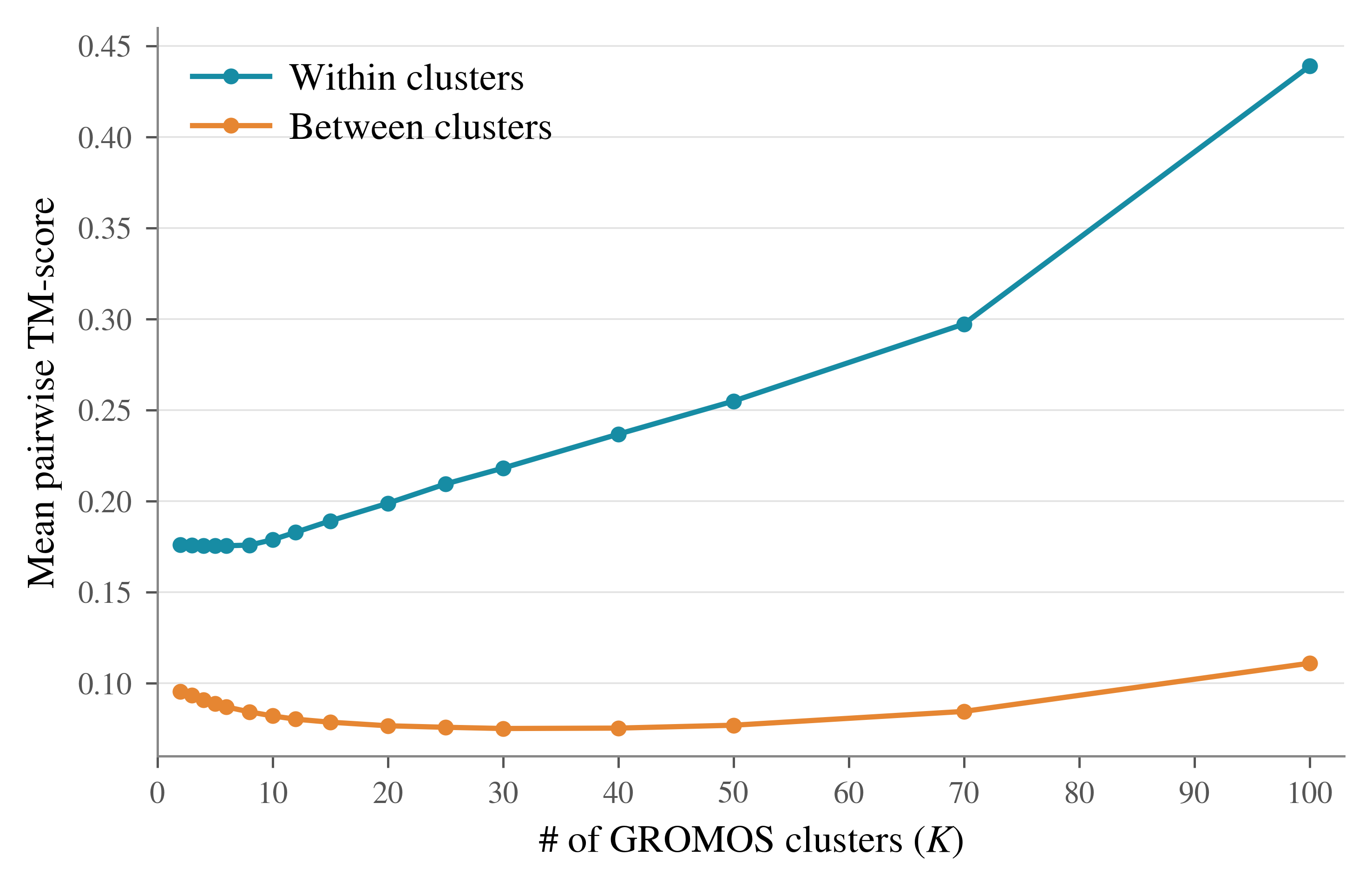}
    \caption{\small Structural separation over conformer bank sizes $K$. Mean within-cluster
    and between-cluster TM-scores.}
    \label{fig:cluster-separation}
\end{figure}

Further, scalar gate and gated attention build on a strong premise that a few members of the bag carry the label while the rest are uninformative, and the fact that it fails to improve on a simple average suggests that identifying and upweighting a small subset of conformers is unnecessary in our setting. To offer potential explanation, we study additional characterizations of the sampled ensembles.
Across 20,386 peptides, conformers from the same peptide are highly diverse: the median mean pairwise TM-score is 0.053, and 88\% of peptides have a mean pairwise score below 0.2 (Table~\ref{tab:diversity}).
This suggests that most peptides' sampled conformers exhibit a continuum instead of interconverting states.
Diversity is particularly high among shorter peptides, which make up much of our benchmark.

We find similarly weak separation when applying the GROMOS clustering procedure used to construct the conformer banks.
Across 1,000 peptides sampled across all length buckets, the mean within-cluster TM-score at $K=10$ is only 0.179.
Moreover, the difference between within- and between-cluster similarity continues to increase as $K$ grows from 2 to 100, without showing an early plateau (Figure~\ref{fig:cluster-separation}).
These observations do not support a scenario where a small number of well-separated conformational groups dominate the sampled ensembles.
Overall, uniformly aggregating representations across the sampled ensemble is at least as effective as learning conformer-specific weights while adding no trainable parameters.
We therefore use mean pooling as the default fusion operator in the remaining experiments.

\begin{table*}[t]
\centering
\caption{\small EASE with different fusion operators, paired difference over same folds against reference conformer (folds improved).}
\label{tab:fusion-paired}
\resizebox{\textwidth}{!}{%
\begin{tabular}{ll|cccccc|c}
\toprule
\textbf{Encoder} & \textbf{Fusion} & \textbf{Toxicity} & \textbf{Hemolytic} & \textbf{HC50} & \textbf{\emph{E. coli} MIC} & \textbf{Stability} & \textbf{Interaction} & \textbf{Mean} \\
\midrule
\multirow{4}{*}{\textbf{GearNet-Edge}}
& Mean
& \textbf{+0.0339 (5/5)} & +0.0116 (5/5) & +0.0321 (5/5) & +0.0196 (5/5) & \textbf{+0.0404 (3/5)} & +0.0267 (5/5) & +0.0274 \\
& Gated
& +0.0331 (5/5) & +0.0133 (5/5) & +0.0345 (5/5) & \textbf{+0.0208 (5/5)} & +0.0396 (3/5) & \textbf{+0.0294 (5/5)} & \textbf{+0.0285} \\
& Attention
& +0.0318 (5/5) & +0.0093 (4/5) & \textbf{+0.0353 (5/5)} & +0.0091 (3/5) & +0.0109 (3/5) & +0.0151 (5/5) & +0.0186 \\
& Gated attn
& +0.0333 (5/5) & \textbf{+0.0136 (5/5)} & +0.0350 (5/5) & +0.0200 (4/5) & +0.0317 (3/5) & +0.0271 (5/5) & +0.0268 \\
\midrule
\multirow{4}{*}{\textbf{ESM-3}}
& Mean
& \textbf{+0.0266 (5/5)} & \textbf{+0.0058 (4/5)} & +0.0238 (4/5) & +0.0327 (5/5) & $-$0.0108 (0/5) & \textbf{+0.0191 (5/5)} & \textbf{+0.0162} \\
& Gated
& +0.0133 (5/5) & +0.0046 (4/5) & +0.0150 (4/5) & \textbf{+0.0372 (5/5)} & \textbf{+0.0075 (4/5)} & +0.0062 (4/5) & +0.0140 \\
& Attention
& +0.0204 (4/5) & +0.0057 (4/5) & +0.0282 (4/5) & +0.0331 (5/5) & $-$0.0043 (1/5) & $-$0.0163 (0/5) & +0.0111 \\
& Gated attn
& +0.0116 (5/5) & +0.0036 (4/5) & \textbf{+0.0330 (5/5)} & +0.0355 (5/5) & $-$0.0217 (1/5) & +0.0026 (3/5) & +0.0108 \\
\bottomrule
\end{tabular}}
\end{table*}


\begin{table}[t]
\centering
\caption{\small Structural diversity of the sampled conformer sets.
Length (Len.) and number of conformers (Conf.) are reported as medians across peptides.
Pairwise TM-scores are computed between conformers of the same peptide
and summarized over each task's sampled corpus.}
\label{tab:diversity}
\small
\setlength{\tabcolsep}{2.5pt}
\begin{tabular}{
    @{}lcc
    @{\hspace{5pt}}c
    @{\hspace{2pt}}c
    @{\hspace{2pt}}c@{}
}
\toprule
\multirow{2}{*}{Task}
& \multirow{2}{*}{Len.}
& \multirow{2}{*}{Conf.}
& \multicolumn{3}{c}{Pairwise TM-score} \\
\cmidrule(lr){4-6}
& & & Median [IQR] & $\%<0.2$ & $\% \geq 0.5$ \\
\midrule
Toxicity
    & 36 & 142 & 0.095 [.055-.307] & 67.0 & 12.8 \\
Hemolytic
    & 20 & 145 & 0.054 [.033-.113] & 86.0 & 3.5 \\
HC50
    & 17 & 146 & 0.052 [.036-.083] & 93.8 & 0.9 \\
\emph{E. coli} MIC
    & 20 & 144 & 0.051 [.033-.086] & 92.5 & 1.0 \\
Stability
    & 21 & 142 & 0.054 [.033-.087] & 89.9 & 0.7 \\
PepPI
    & 11 & 145 & 0.041 [.027-.061] & 99.1 & 0.1 \\
\bottomrule
\end{tabular}
\end{table}

\subsection{Performance improves with bank size}
\label{sec:res-k}

\begin{figure*}[t]
\centering
\includegraphics[width=\linewidth]{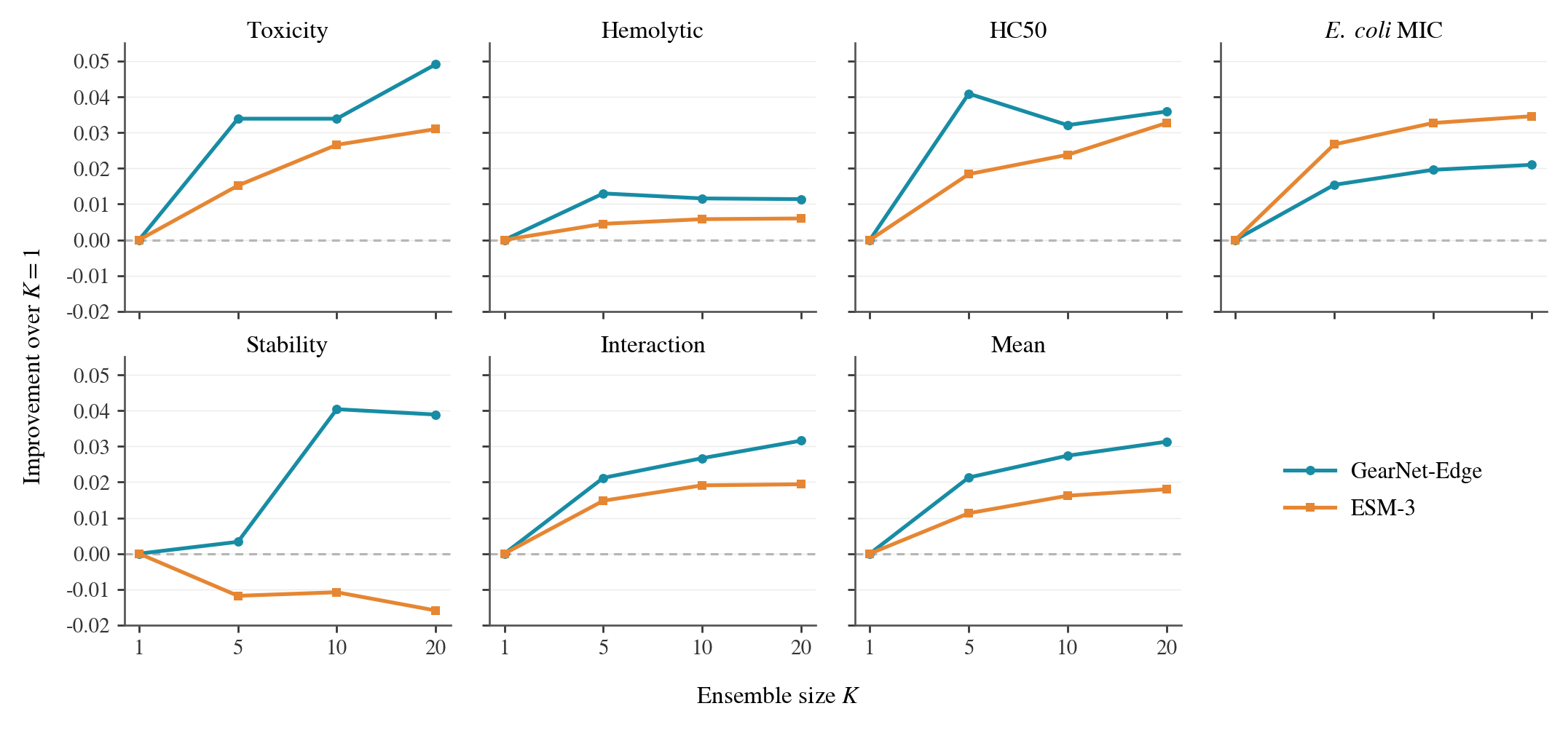}
\caption{\small Effect of ensemble size on task performance.
Each panel reports the paired difference with the reference conformer on the same folds. Positive values indicates improvement for both classification (AUROC) and regression (RMSE) tasks.}
\label{fig:ensemble-trends}
\end{figure*}

In this section, we report the ablation study on the conformer bank size $K$, varying $K$ over 5, 10, 20. 
Figure~\ref{fig:ensemble-trends} shows how performance changes with ensemble size relative to the reference conformer baseline (treated as $K$=1). Absolute and paired performance (Tables~\ref{tab:ksweep-abs},~\ref{tab:ksweep-paired}) are given in the Appendix. Averaged across tasks, both encoders benefit increasingly from larger ensembles through $K=20$, with mean improvements rising from $+0.0213$ at $K=5$ to $+0.0274$ at $K=10$ and $+0.0313$ at $K=20$ for GearNet-Edge and from $+0.0113$ at $K=5$ to $+0.0162$ at $K=10$ and $+0.0180$ at $K=20$ for ESM-3. The task-level trends are not uniformly monotonic, as some gains saturate or fluctuate and ESM-3 does not improve blood stability. Nevertheless, the overall trend shows that the benefits of EASE increase with diminishing returns as additional conformers are included.

This trend is consistent with the structural diversity measured in the sampled ensemble study in 
\S\ref{sec:res-fusion}.
As $K$ increases, the selected conformers continue to cover structurally distinct regions of the sampled ensemble, while the marginal benefit of each additional conformer decreases.
In this sense, $K$ acts primarily as a sampling budget, as larger $K$ values provide broader ensemble coverage at proportionally greater computational cost.
This trade-off motivates that we use $K=10$ in the remaining experiments.
It achieves approximately 88\% of GearNet-Edge's average improvement observed at $K=20$ and 90\% of ESM-3's, while requiring half as many conformer encodings.

\subsection{Alternative ways of using the ensemble}
\label{sec:res-alt}

In addition to EASE, we compare six alternative mechanisms on GearNet-Edge, using the same conformational ensemble, data splits, and training protocol. 
The alternative methods incorporate ensemble summary statistics through graph features, edge geometry, topology, intermediate layers, the readout, or auxiliary supervision. Full definitions are provided in Appendix~\ref{sec:alternatives}.




\smallskip
\noindent\textbf{Alternative mechanisms comparison.}
Table~\ref{tab:alt-paired} in the Appendix reports the paired differences over same folds against each method's baseline. 
Across all methods of injecting ensemble-derived statistics into the model, average improvements over the base encoder span a fivefold range, from $+0.0028$ to $+0.0154$.
The two mechanisms that make the ensemble statistics act throughout the encoder perform best: stochastic topology (3) improves by $+0.0154$, and layer-wise modulation (4) by $+0.0130$ with residue statistics and $+0.0117$ with residue and pair statistics.
The mechanisms that inject a fixed summary at a single point tend to perform worse: node and edge features (1) improve by $+0.0067$ and $+0.0064$, the pooled descriptor at the readout (5) by $+0.0057$, and auxiliary supervision (6) by $+0.0036$ and $+0.0028$.
These results show that the injection point matters among the summary-statistics mechanisms.

\noindent\textbf{Q: Does increasing the detail of the summary statistics help? A: No.}
Replacing the first two moments of the pairwise distance distribution with a 16-bin distributional representation (2) improves over the base encoder by $+0.0055$, slightly below the $+0.0064$ for the pairwise moments it replaces (1).
Adding pairwise statistics on top of residue statistics similarly changes average performance by $-0.0003$.
Within these tests, retaining additional pairwise or distributional detail does not recover the advantage of encoding the conformers directly.

\noindent\textbf{Q: Direct conformer encoding vs. ensemble summary statistics? A: Direct conformer encoding.}
EASE improves over its reference-conformer baseline by $+0.0274$
on average, compared with $+0.0028$ to $+0.0154$ for the alternative summary-statistics approaches over their base encoder.
These results favor preserving entire conformers as structural inputs rather than first compressing the ensemble to predefined statistics.
One possible explanation is that summary statistics such as fluctuation, contact frequency, and distance moments marginalize over the ensemble before structural encoding and might abstract away spatial arrangement of residues within each structure. On the other hand, EASE preserves such joint geometry of each sampled structure for the pretrained encoder to process.
We therefore conclude that the summary statistics tested here do not recover the information available from directly encoding the conformers.

\subsection{When Does Ensemble Information Complement Protein Language Models?}
\label{sec:res-seq}

\noindent\textbf{Comparison with PLM baselines.}
We first investigate whether conformational ensembles provide additional predictive information beyond pretrained protein language model (PLM) representations. Table~\ref{tab:sequence} compares EASE and its reference conformer baselines against sequence-only PLM representations.
ESM-3 provides the most controlled comparison because its PLM baseline uses the same pretrained backbone, adapters, and training schedule, only with the structure track masked.
The matched reference conformer underperforms sequence-only ESM-3 by $0.0058$ on average and performs better on three of six tasks.
EASE reverses this pattern, improving over sequence-only ESM-3 by $+0.0104$ on average and on five of six tasks.
The only exception is HC50, where sequence-only ESM-3 is better by $0.0014$. Thus, although adding a single structure provides no consistent advantage over the PLM track alone, conditioning the same model on a conformational ensemble using EASE can provide complementary predictive information across most tasks.
On the other hand, since GearNet-Edge has no sequence channel on its own, we instead compare it with ESM-2-35M, a separately pretrained PLM of comparable size using the same downstream adaptation protocol.
Unlike the ESM-3 comparison, differences here can reflect both input modality and pretrained encoder.
The reference conformer on GearNet-Edge generally underperforms ESM-2, whereas EASE narrows this gap and outperforms ESM-2 on three of six tasks.
The largest advantages occur on blood stability and peptide-protein interaction, where conformational information matters biologically: proteolytic stability depends in part on the accessibility and local geometry of potential cleavage sites, and protein binding depends on the availability of conformations compatible with the receptor protein surface.
As a result, although EASE on GearNet-Edge does not consistently outperform the PLM baseline, conditioning the structure encoder on an ensemble provides competitive predictive information without relying on pretrained language representations.


\smallskip
\noindent\textbf{Encoder-dependent ensemble gains and conformational variation.}
To further investigate how the integration of PLM representations influence the benefit of ensemble conditioning, we extend our evaluation to ESM-GearNet, which incorporates pretrained ESM-2 representations into a geometric graph encoder. EASE achieves an average paired improvement of +0.0077 across six tasks on ESM-GearNet, which is substantially smaller than +0.0162 on ESM-3 and +0.0274 on GearNet-Edge (Tables~\ref{tab:esmgearnet-abs}, ~\ref{tab:esmgearnet-paired}). To examine whether these differences are related to how the encoders represent conformational variation, we measure the proportion of embedding variation attributable to differences between conformers of the same peptide. For each encoder and task, we compute \(\eta^2 = SS_{\mathrm{within}}/(SS_{\mathrm{within}}+SS_{\mathrm{between}})\), where \(SS_{\mathrm{within}}\) is the variation across conformers of the same peptide and \(SS_{\mathrm{between}}\) is the variation between different peptides. Figure~\ref{fig:eta-vs-gain} shows the relationship between \(\eta^2\) and the paired improvement of EASE over the matched reference conformer. GearNet-Edge exhibits the highest mean \(\eta^2\) (0.562), followed by ESM-3 (0.330) and ESM-GearNet (0.164), and this ordering matches that of their average EASE gains. This suggests that encoders with greater conformer variation in their representations may benefit more from aggregating multiple conformations.
One possible explanation for this trend is how PLM information enters each encoder. GearNet-Edge relies solely on geometric message passing without pretrained PLM components, which allows structural differences to directly influence its representations. ESM-3 jointly processes sequence and structure tokens through transformer layers, potentially reducing the relative contribution of conformer structure information. ESM-GearNet incorporates PLM information in two places: ESM-2 embeddings are both passed as node features for its graphs and concatenated directly to the final representation. This high PLM involvement may further reduce the influence of conformational differences, which is consistent with its lowest $\eta^2$ and smallest ensemble gains. 

\section{Conclusion}
\label{sec:conclusion}

We introduced EASE, a framework for extracting predictive representations from peptide conformational ensembles.
EASE improves over single-structure baselines on all six tasks on GearNet-Edge and five on ESM-3.
Parameter-free mean pooling is competitive with learned fusion, and larger conformer banks provide diminishing gains through $K=20$.
Six alternative approaches that first reduce the ensemble to summary statistics yield smaller improvements over their respective baselines, supporting the value of encoding whole conformers directly instead of compressing ensemble information into dynamics descriptors.
Comparisons with sequence-only PLMs further show that EASE provides predictive information beyond sequence representations. Across three encoders, greater conformer embedding variation is associated with larger ensemble gains, suggesting that how PLM information is integrated into structural representations influences the benefit of ensemble conditioning.

\section{Limitations}
\label{sec:limitations}

There are several limitations to our current work. First, this study focuses on improving peptide structural representation learning but does not claim that structure is the strongest modality for peptide property prediction.
Sequence-only baselines remain stronger on half of the six downstream tasks evaluated on GearNet-Edge (\S\ref{sec:res-seq}), and EASE only narrows this gap.
Second, our experiments are restricted to linear peptides composed of canonical amino acids. This restriction excludes cyclic and non-canonical peptides, which constitute an important part of therapeutic peptide space but are not supported by BioEmu.
Further, the ensembles generated by BioEmu approximate equilibrium conformational distributions instead of providing calibrated Boltzmann samples or molecular-dynamics trajectories. Our results therefore concern the use of multiple plausible conformations, without modeling how frequently each state occurs or how the peptide transitions between them.
Finally, our work has potential ethical and environmental implications. Generating and encoding multiple conformers increases computational cost and associated energy consumption compared with single-structure approaches. Additionally, the predicted properties should not be treated as substitutes for experimental validation, particularly in safety-critical applications.

\FloatBarrier
\bibliography{references}

\clearpage
\appendix

\section{Fusion operators details}

Table~\ref{tab:fusion-params} reports the number of parameters introduced by each fusion operator for both encoders.

\begin{table}[t]
\centering
\caption{Parameters added by each fusion operator.}
\label{tab:fusion-params}
\small
\setlength{\tabcolsep}{2.5pt}
\resizebox{\columnwidth}{!}{%
\begin{tabular}{@{}llcc@{}}
\toprule
\textbf{Encoder} & \textbf{Fusion} & \textbf{Internal width} & \textbf{Params added} \\
\midrule
GearNet-Edge ($d=512$)
 & mean            & 512 & 0 \\
 & gated           & 512 & 513 \\
 & gated attn & 512 & 131,457 \\
 & attention       & 64  & 99,584 \\
\midrule
ESM-3 ($d=1536$)
 & mean            & 1536 & 0 \\
 & gated           & 1536 & 1,537 \\
 & gated attn & 1536 & 393,601 \\
 & attention       & 128  & 132,480 \\
\bottomrule
\end{tabular}%
}
\end{table}

\section{Alternative methods details}

\subsection{Alternative uses of conformational ensembles}
\label{sec:alternatives}

EASE preserves each sampled conformer as a structural input and aggregates their encoded representations. To test whether the same ensemble information can be used more effectively elsewhere in the model, we construct a set of six alternative mechanisms that expose the encoder to summary statistics computed from the ensemble. All alternatives draw on the same BioEmu-generated conformational ensembles. Since multiple alternative mechanisms apply to geometric graph features, we evaluate these mechanisms on GearNet-Edge only.


\begin{table*}[t]
\centering
\caption{Ensemble statistics available to the alternative mechanisms. Residue and pairwise quantities enter the encoder as node features, edge features or an added relation; the peptide-level descriptor enters at the readout. The form column indicates whether a quantity is injected in raw form, within-peptide z-scored form, or both.}
\label{tab:ensemble-stats}
\small
\begin{tabularx}{\textwidth}{@{}lllX@{}}
\toprule
\textbf{Level} & \textbf{Quantity} & \textbf{Form} & \textbf{Definition} \\
\midrule
residue & fluctuation & both & RMS deviation of the C$\alpha$ position around the ensemble mean \\
residue & accessibility, mean and s.d. & both & per-residue solvent-accessible surface area across conformers \\
residue & dihedral variability & raw & circular variance of backbone $\phi$ and $\psi$ across conformers \\
pair & distance, mean and s.d. & both & C$\alpha$--C$\alpha$ distance across conformers \\
pair & contact frequency & raw & fraction of conformers with C$\alpha$--C$\alpha$ separation below 10 \AA{} \\
pair & ensemble-contact relation & --- & additional edge type for pairs with contact frequency $\geq 0.3$ \\
peptide & pooled descriptor & both & 22-dimensional summary of fluctuation, accessibility, distance, radius of
  gyration, dihedral variability, contact statistics, and peptide length \\
\bottomrule
\end{tabularx}
\end{table*}

\smallskip
\noindent\textbf{Ensemble summary statistics.}
The six alternatives first compact the sampled conformers into residue-level, pair-level, or peptide-level statistics. At the residue level, we compute the RMS fluctuation of each C$\alpha$ position around its ensemble mean, the mean and standard deviation of per-residue solvent accessibility, and the circular variance of the backbone $\phi$ and $\psi$ dihedral angles. At the pair level, we compute the mean and standard deviation of C$\alpha$--C$\alpha$ distance and the contact frequency, defined as the fraction of conformers in which the pair is separated by less than 10 \AA{}. Some of these quantities are injected in both raw and within-peptide z-scored form, as the raw values preserve magnitude across peptides while the z-scored values retain only the within-peptide profile. A summary of all statistics is listed in Table~\ref{tab:ensemble-stats}. 

\smallskip
\noindent\textbf{(1) Adding graph features and edges.}
We first inject ensemble statistics directly into GearNet-Edge.
Residue-level statistics are concatenated to the node features, while pairwise statistics are concatenated to edge features.
The pairwise variant additionally introduces an ensemble-contact edge type, defined as residue pairs whose contact frequency is at least 0.3.

\smallskip
\noindent\textbf{(2) Distributional edge geometry.}
The pairwise features above summarize the distance between each residue pair using its mean and standard deviation.
These moments cannot distinguish distributions with similar mean and variance but different shapes.
We therefore replace the distance moments with a 16-bin soft histogram of the full distance distribution, obtained by averaging Gaussian radial basis functions centered between 2 and 32 \AA{} across conformers.
This representation preserves absolute distance scale while retaining more information about the distribution of distances across the ensemble.
Graph topology is unchanged and only the features carried by each edge are replaced.

\smallskip
\noindent\textbf{(3) Stochastic ensemble topology.}
The ensemble-contact edge type above converts contact frequency into a binary edge type using a fixed threshold, so a residue pair either is or is not in contact.
In this mechanism, we instead use contact frequency directly to define a distribution over graph topology.
During training, each non-sequential edge is retained independently with probability equal to its contact frequency, while sequential backbone edges are always retained.
For instance, a pair in contact in half the conformers contributes to half the training steps.
At evaluation, we use the expectation by weighting each edge's message by its contact frequency.
This introduces ensemble-dependent topology without additional trainable parameters or a contact-frequency threshold.

\smallskip
\noindent\textbf{(4) Layer-wise feature modulation.}
In the mechanisms above, the statistics enter the encoder only once, at the input layer.
Thus, these statistics survive only as what the hidden state retains through six rounds of relational convolution, normalization and residual addition.
We instead make them available throughout the encoder using feature-wise linear modulation (FiLM) \citep{perez2018film}.
At each layer, a linear map transforms the residue-level statistics into a feature-wise scale and shift, and updates the hidden state as $h_l \leftarrow \gamma_l \odot h_l + \beta_l$. Each layer receives the original ensemble statistics rather than the output of the preceding modulation.
The maps are initialized so that $\gamma_l=1$ and $\beta_l=0$, preserving the pretrained function at initialization.
Modulation is applied independently at each residue, and the hidden dimensionality is unchanged.

\smallskip
\noindent\textbf{(5) Peptide-level descriptor at the readout.}
The mechanisms above provide per-residue or pairwise quantities, so no quantity at the entire peptide scope is used in the model.
In this mechanism, we pool the ensemble statistics into a 22-dimensional peptide descriptor and concatenate it to the pooled peptide representation at the readout.
The descriptor summarizes fluctuation, solvent accessibility, pairwise distance, radius of gyration, dihedral variability, contact frequency, long-range contacts, contact order, transient contacts, and peptide length.
It includes both the raw and within-peptide normalized summaries where applicable.
Descriptor dimensions are standardized using statistics from the training fold.
The encoder itself is unchanged.

\smallskip
\noindent\textbf{(6) Dynamics as auxiliary supervision.}
We next use ensemble statistics as training targets rather than model inputs.
Auxiliary heads on the final residue representations predict three per-residue quantities: RMS fluctuation and the mean and standard deviation of solvent accessibility.
A second variant additionally predicts pairwise contact probability from a down-projected inner product of residue representations.
The training objective is $\mathcal{L} = \mathcal{L}{\mathrm{task}} + \lambda \mathcal{L}{\mathrm{aux}}$, with a second weighted term for the pairwise head in the latter variant.
The auxiliary heads are discarded at inference, so the ensemble is used only during training.
Rather than tuning $\lambda$ as a task-specific hyperparameter, we calibrate it so that the task and auxiliary objectives contribute gradients of equal norm to the shared trunk, measured over training batches.

\smallskip
\noindent\textbf{Baseline for alternative methods.}
The alternative mechanisms are compared against the unmodified GearNet-Edge encoder given a single structure, chosen as the conformer whose C$\alpha$ coordinates lie closest to the ensemble mean.
We note that this is a different structure from the reference conformer of \S\ref{sec:baselines}, which is drawn from the $K$ = 10 conformer bank.
The reference conformer baseline remains specific to EASE, where the contrast has to isolate fusing the bank from selecting a member of it; here the encoder input is a single conformer in every arm, and what varies is the additional ensemble information supplied alongside it.

\noindent\textbf{Evaluation of alternative methods.} Each of these alternative methods is evaluated at five folds and four training seeds on all six tasks. 

\section{Training protocol details}
\label{sec:training-config}

\noindent\textbf{Configurations and hyperparameters.} Tables~\ref{tab:training} and ~\ref{tab:selected-hparams} show the specific training configurations, hyperparameter search grids, and the selected hyperparameters.

\begin{table*}[t]
\centering
\caption{Training configuration for each encoder. }
\label{tab:training}
\small
\begin{tabularx}{\textwidth}{@{}l
    >{\centering\arraybackslash}X
    >{\centering\arraybackslash}X@{}}
\toprule
Setting & GearNet-Edge & ESM-3 \\
\midrule
Optimizer & AdamW & AdamW \\
Trunk learning rate & $3\times10^{-4}$ & tuned, $\{10^{-4},10^{-5}\}$ \\
Head learning rate & $3\times10^{-4}$ & $10^{-3}$ \\
Weight decay & $0.1$ & $0.01$ \\
Batch size & 16 & tuned, $\{16,32\}$ \\
LoRA rank & tuned, $\{4,8\}$ & 8 \\
Head dropout & tuned, $\{0.1,0.3,0.5\}$ & 0.1 \\
Max epochs & 100 & 20 \\
Early-stopping patience & 15 & 6 \\
PPI batch size & 64 & tuned, $\{16,32\}$ \\
PPI max epochs & 120 & 50 \\
PPI patience & 15 & 12 \\
Tuning granularity & one config per task, shared by all arms & one config per (task, arm) \\
Seeds / fold & 4 & 3 \\
Head hidden width & 64 & 128 \\
Trainable parameters & 0.46--0.63M & 3.77--4.27M \\
\bottomrule
\end{tabularx}
\end{table*}

\begin{table*}[t]
\centering
\caption{Selected configurations. GearNet-Edge, the ESM-2-35M sequence-only baseline, and the ChemBERTa chemical baseline each tune one configuration per task over LoRA rank $\{4,8\}$ and head dropout $\{0.1,0.3,0.5\}$. ESM-3 tunes one configuration per (task, arm) over trunk learning rate $\{10^{-4},10^{-5}\}$ and batch size $\{16,32\}$.}
\label{tab:selected-hparams}
\resizebox{\textwidth}{!}{%
\begin{tabular}{@{}lcccccccccccc@{}}
\toprule
 & \multicolumn{2}{c}{GearNet-Edge} & \multicolumn{2}{c}{ESM-2 (35M)} & \multicolumn{2}{c}{ChemBERTa} & \multicolumn{6}{c}{ESM-3 (trunk learning rate / batch size)} \\
\cmidrule(lr){2-3}\cmidrule(lr){4-5}\cmidrule(lr){6-7}\cmidrule(lr){8-13}
task & LoRA rank & head dropout & LoRA rank & head dropout & LoRA rank & head dropout & sequence & reference conformer & mean & gated & gated attn. & attention \\
\midrule
toxicity            & 8 & 0.5 & 8 & 0.1 & 8 & 0.5 & $10^{-5}$/16 & $10^{-4}$/32 & $10^{-4}$/16 & $10^{-4}$/32 & $10^{-4}$/16 & $10^{-4}$/32 \\
hemolytic           & 8 & 0.1 & 8 & 0.1 & 8 & 0.3 & $10^{-4}$/16 & $10^{-4}$/16 & $10^{-4}$/16 & $10^{-4}$/16 & $10^{-4}$/16 & $10^{-4}$/16 \\
hemolytic HC50      & 8 & 0.1 & 8 & 0.5 & 4 & 0.3 & $10^{-4}$/16 & $10^{-4}$/16 & $10^{-4}$/32 & $10^{-4}$/32 & $10^{-4}$/16 & $10^{-4}$/16 \\
\emph{E. coli} MIC  & 8 & 0.5 & 8 & 0.5 & 4 & 0.5 & $10^{-4}$/16 & $10^{-5}$/16 & $10^{-4}$/16 & $10^{-4}$/32 & $10^{-4}$/16 & $10^{-4}$/32 \\
blood stability     & 8 & 0.5 & 4 & 0.1 & 8 & 0.5 & $10^{-5}$/16 & $10^{-5}$/16 & $10^{-5}$/32 & $10^{-5}$/16 & $10^{-4}$/16 & $10^{-5}$/16 \\
interaction         & 8 & 0.3 & 4 & 0.1 & 4 & 0.1 & $10^{-5}$/32 & $10^{-5}$/32 & $10^{-5}$/32 & $10^{-5}$/32 & $10^{-5}$/32 & $10^{-5}$/32 \\
\bottomrule
\end{tabular}}
\end{table*}

\smallskip
\noindent\textbf{Computational resources.} Every training run used a single GPU, and the GPU is randomly scheduled across NVIDIA RTX 6000 Ada (48 GB), A100 (80 GB), and RTX PRO 6000 Blackwell (96 GB). 
The total computational budget was approximately 2,200 GPU-hours and 2,000 CPU-hours. We spent approximately 100 GPU-hours for BioEmu conformational ensemble generation (around 31.6 k peptides and 3.4 M conformers in total), including AlphaFold2 embedding and diffusion sampling. An additional 2,000 CPU-hours was required for the all-atom side-chain reconstruction and energy minimization. 
On the model training and evaluation part, we spent approximately 2,070 GPU-hours on the reported experiments involving EASE and its ablations, along with all baselines. 

\smallskip
\noindent\textbf{Models and trainable parameters.}
ESM-3 (esm3\_sm\_open\_v1) contains 1.40 B parameters and GearNet-Edge contains 20.2 M. The ESM-GearNet includes a frozen ESM-2-650M (649 M) with a GearNet trunk (15.7 M). In every experiment, only the LoRA adapters, the fusion layer, and the prediction MLP heads are trained, which is around 3.8 M trainable parameters for ESM-3, 0.46 M for GearNet-Edge, and 0.88 M for ESM-GearNet.

\section{Results: additional comparison tables and figures}

Table ~\ref{tab:fusion-abs} shows the absolute test performance of EASE with each fusion operator.

\begin{table*}[t]
\centering
\caption{\small Test performance of EASE with different fusion operators.
Mean $\pm$ standard deviation over folds with training seeds averaged within each fold.
AUROC is reported for classification and RMSE for regression.
Average rank is obtained by ranking the five arms on each task and averaging
their ranks across the six tasks.
Bold indicates the best mean performance for each task within each encoder.}
\label{tab:fusion-abs}
\resizebox{\textwidth}{!}{%
\begin{tabular}{llcccccc|c}
\toprule
\textbf{Encoder}
& \textbf{Fusion}
& \textbf{Toxicity} $\uparrow$
& \textbf{Hemolytic} $\uparrow$
& \textbf{HC50} $\downarrow$
& \textbf{\emph{E. coli} MIC} $\downarrow$
& \textbf{Stability} $\uparrow$
& \textbf{Interaction} $\uparrow$
& \textbf{Avg. rank} $\downarrow$ \\
\midrule
\multirow{5}{*}{\textbf{GearNet-Edge}}
& Ref. conformer
& 0.6893 $\pm$ 0.1008
& 0.8617 $\pm$ 0.0221
& 0.6460 $\pm$ 0.0191
& 0.6847 $\pm$ 0.0202
& 0.6737 $\pm$ 0.0458
& 0.8059 $\pm$ 0.0159
& 5.00 \\
& Mean
& \textbf{0.7231 $\pm$ 0.0938}
& 0.8733 $\pm$ 0.0225
& 0.6139 $\pm$ 0.0164
& 0.6651 $\pm$ 0.0275
& \textbf{0.7141 $\pm$ 0.0525}
& 0.8327 $\pm$ 0.0101
& 2.50 \\
& Gated
& 0.7224 $\pm$ 0.0952
& 0.8750 $\pm$ 0.0218
& 0.6115 $\pm$ 0.0182
& \textbf{0.6639 $\pm$ 0.0313}
& 0.7134 $\pm$ 0.0413
& \textbf{0.8353 $\pm$ 0.0097}
& \textbf{2.00} \\
& Attention
& 0.7210 $\pm$ 0.0939
& 0.8710 $\pm$ 0.0208
& \textbf{0.6107 $\pm$ 0.0217}
& 0.6755 $\pm$ 0.0232
& 0.6846 $\pm$ 0.0723
& 0.8210 $\pm$ 0.0110
& 3.50 \\
& Gated attn
& 0.7226 $\pm$ 0.0956
& \textbf{0.8753 $\pm$ 0.0217}
& 0.6110 $\pm$ 0.0191
& 0.6647 $\pm$ 0.0313
& 0.7055 $\pm$ 0.0495
& 0.8331 $\pm$ 0.0088
& \textbf{2.00} \\
\midrule
\multirow{5}{*}{\textbf{ESM-3}}
& Ref. conformer
& 0.7792 $\pm$ 0.0534
& 0.9013 $\pm$ 0.0209
& 0.5793 $\pm$ 0.0230
& 0.6842 $\pm$ 0.0225
& 0.7141 $\pm$ 0.0266
& 0.8329 $\pm$ 0.0118
& 4.33 \\
& Mean
& \textbf{0.8058 $\pm$ 0.0387}
& \textbf{0.9071 $\pm$ 0.0161}
& 0.5554 $\pm$ 0.0334
& 0.6515 $\pm$ 0.0183
& 0.7032 $\pm$ 0.0314
& \textbf{0.8520 $\pm$ 0.0100}
& \textbf{2.33} \\
& Gated
& 0.7925 $\pm$ 0.0515
& 0.9059 $\pm$ 0.0192
& 0.5642 $\pm$ 0.0545
& \textbf{0.6470 $\pm$ 0.0189}
& \textbf{0.7216 $\pm$ 0.0286}
& 0.8391 $\pm$ 0.0098
& \textbf{2.33} \\
& Attention
& 0.7996 $\pm$ 0.0258
& 0.9070 $\pm$ 0.0171
& 0.5511 $\pm$ 0.0423
& 0.6511 $\pm$ 0.0124
& 0.7098 $\pm$ 0.0133
& 0.8166 $\pm$ 0.0090
& 2.83 \\
& Gated attn
& 0.7908 $\pm$ 0.0514
& 0.9049 $\pm$ 0.0196
& \textbf{0.5462 $\pm$ 0.0318}
& 0.6488 $\pm$ 0.0164
& 0.6924 $\pm$ 0.0163
& 0.8355 $\pm$ 0.0079
& 3.17 \\
\bottomrule
\end{tabular}}
\end{table*}

Tables~\ref{tab:ksweep-abs} and ~\ref{tab:ksweep-paired} displays the absolute test performance and paired difference per fold against the reference conformer of the EASE method as the ensemble size $K$ increases through {5, 10, 20}.

\begin{table*}[t]
\centering
\caption{Ensemble sizes $K$, absolute test performance. Mean $\pm$ standard deviation over folds with training seeds averaged within each fold. Mean fusion is used for every ensemble size. AUROC reported for classification and RMSE for regression.}
\label{tab:ksweep-abs}
\resizebox{\textwidth}{!}{%
\begin{tabular}{llcccccc}
\toprule
encoder & arm & toxicity $\uparrow$ & hemolytic $\uparrow$ & HC50 $\downarrow$ & \emph{E. coli} MIC $\downarrow$ & stability $\uparrow$ & interaction $\uparrow$ \\
\midrule
GearNet-Edge & reference conformer & 0.6893 $\pm$ 0.1008 & 0.8617 $\pm$ 0.0221 & 0.6460 $\pm$ 0.0191 & 0.6847 $\pm$ 0.0202 & 0.6737 $\pm$ 0.0458 & 0.8059 $\pm$ 0.0159 \\
GearNet-Edge & $K{=}5$ & 0.7231 $\pm$ 0.0953 & \textbf{0.8747 $\pm$ 0.0234} & \textbf{0.6051 $\pm$ 0.0231} & 0.6692 $\pm$ 0.0249 & 0.6770 $\pm$ 0.0496 & 0.8271 $\pm$ 0.0108 \\
GearNet-Edge & $K{=}10$ & 0.7231 $\pm$ 0.0938 & 0.8733 $\pm$ 0.0225 & 0.6139 $\pm$ 0.0164 & 0.6651 $\pm$ 0.0275 & \textbf{0.7141 $\pm$ 0.0525} & 0.8327 $\pm$ 0.0101 \\
GearNet-Edge & $K{=}20$ & \textbf{0.7384 $\pm$ 0.0876} & 0.8731 $\pm$ 0.0197 & 0.6101 $\pm$ 0.0256 & \textbf{0.6636 $\pm$ 0.0248} & 0.7126 $\pm$ 0.0604 & \textbf{0.8376 $\pm$ 0.0137} \\
\midrule
ESM-3 & reference conformer & 0.7792 $\pm$ 0.0534 & 0.9013 $\pm$ 0.0209 & 0.5793 $\pm$ 0.0230 & 0.6842 $\pm$ 0.0225 & \textbf{0.7141 $\pm$ 0.0266} & 0.8329 $\pm$ 0.0118 \\
ESM-3 & $K{=}5$ & 0.7944 $\pm$ 0.0498 & 0.9058 $\pm$ 0.0161 & 0.5608 $\pm$ 0.0400 & 0.6575 $\pm$ 0.0217 & 0.7022 $\pm$ 0.0525 & 0.8477 $\pm$ 0.0116 \\
ESM-3 & $K{=}10$ & 0.8058 $\pm$ 0.0387 & 0.9071 $\pm$ 0.0161 & 0.5554 $\pm$ 0.0334 & 0.6515 $\pm$ 0.0183 & 0.7032 $\pm$ 0.0314 & 0.8520 $\pm$ 0.0100 \\
ESM-3 & $K{=}20$ & \textbf{0.8102 $\pm$ 0.0318} & \textbf{0.9073 $\pm$ 0.0193} & \textbf{0.5466 $\pm$ 0.0458} & \textbf{0.6496 $\pm$ 0.0186} & 0.6982 $\pm$ 0.0262 & \textbf{0.8523 $\pm$ 0.0102} \\
\bottomrule
\end{tabular}}
\end{table*}

\begin{table*}[t]
\centering
\caption{Ensemble sizes $K$, paired differences over same folds against the fixed reference-conformer baseline (folds improved out of five). Positive values indicate improvement.}
\label{tab:ksweep-paired}
\resizebox{\textwidth}{!}{%
\begin{tabular}{lccccccccc}
\toprule
encoder & $K$ & toxicity & hemolytic & HC50 & \emph{E. coli} MIC & stability & interaction & mean & tasks $\uparrow$ \\
\midrule
GearNet-Edge & 5 & +0.0339 (5/5) & +0.0130 (5/5) & +0.0409 (5/5) & +0.0154 (4/5) & +0.0033 (2/5) & +0.0212 (5/5) & +0.0213 & 6/6 \\
GearNet-Edge & 10 & +0.0339 (5/5) & +0.0116 (5/5) & +0.0321 (5/5) & +0.0196 (5/5) & +0.0404 (3/5) & +0.0267 (5/5) & +0.0274 & 6/6 \\
GearNet-Edge & 20 & +0.0491 (5/5) & +0.0114 (5/5) & +0.0359 (5/5) & +0.0210 (5/5) & +0.0389 (3/5) & +0.0316 (5/5) & \textbf{+0.0313} & 6/6 \\
\midrule
ESM-3 & 5 & +0.0152 (5/5) & +0.0045 (4/5) & +0.0184 (4/5) & +0.0267 (5/5) & $-$0.0118 (2/5) & +0.0148 (5/5) & +0.0113 & 5/6 \\
ESM-3 & 10 & +0.0266 (5/5) & +0.0058 (4/5) & +0.0238 (4/5) & +0.0327 (5/5) & $-$0.0108 (0/5) & +0.0191 (5/5) & +0.0162 & 5/6 \\
ESM-3 & 20 & +0.0310 (5/5) & +0.0060 (4/5) & +0.0327 (4/5) & +0.0346 (5/5) & $-$0.0159 (0/5) & +0.0194 (5/5) & \textbf{+0.0180} & 5/6 \\
\bottomrule
\end{tabular}}
\end{table*}

The following table~\ref{tab:alt-paired} shows the paired performance improvements over same folds of each alternative method against its own single-structure baseline, as defined in \S\ref{sec:alternatives}. 

\begin{table*}[t]
\centering
\caption{\small Alternative uses of the conformational ensemble. GearNet-Edge, mean paired difference against each mechanism's own single-structure baseline over the same folds (folds improved out of five). The alternative mechanisms are paired against the base encoder with the mean structure and EASE is paired against the reference conformer. Positive is better, with mechanisms numbered as in \S\ref{sec:alternatives}. $^{\dagger}$This row alone is an increment over the row above rather than a difference against a baseline.}
\label{tab:alt-paired}
\resizebox{\textwidth}{!}{%
\begin{tabular}{cl|cccccc|cc}
\toprule
\textbf{\#} & \textbf{Mechanism} & \textbf{Toxicity} & \textbf{Hemolytic} & \textbf{HC50} & \textbf{\emph{E. coli} MIC} & \textbf{Stability} & \textbf{PepPI} & \textbf{Mean} & \textbf{Tasks $\uparrow$} \\
\midrule
1 & Per-residue statistics & +0.0083 (4/5) & +0.0054 (4/5) & +0.0116 (3/5) & $-$0.0013 (3/5) & +0.0109 (5/5) & +0.0056 (4/5) & +0.0067 & 5/6 \\
1 & Pairwise statistics & +0.0127 (4/5) & +0.0068 (3/5) & +0.0044 (3/5) & $-$0.0047 (2/5) & +0.0155 (5/5) & +0.0037 (4/5) & +0.0064 & 5/6 \\
1 & $\hookrightarrow$ Extra edge type$^{\dagger}$ & +0.0044 (2/5) & +0.0014 (3/5) & $-$0.0071 (2/5) & $-$0.0034 (1/5) & +0.0046 (2/5) & $-$0.0019 (2/5) & $-$0.0003 & 3/6 \\
2 & Edge distance histogram & +0.0095 (3/5) & +0.0059 (4/5) & +0.0029 (3/5) & $-$0.0039 (2/5) & +0.0159 (4/5) & +0.0027 (4/5) & +0.0055 & 5/6 \\
3 & Contact-frequency topology & +0.0055 (4/5) & +0.0032 (2/5) & +0.0260 (5/5) & $-$0.0025 (1/5) & \textbf{+0.0499 (4/5)} & +0.0102 (5/5) & +0.0154 & 5/6 \\
4 & Modulation, node stats & +0.0257 (5/5) & +0.0077 (4/5) & +0.0184 (4/5) & $-$0.0024 (3/5) & +0.0206 (4/5) & +0.0081 (5/5) & +0.0130 & 5/6 \\
4 & Modulation, node + edge stats & +0.0265 (5/5) & +0.0043 (3/5) & +0.0188 (4/5) & $-$0.0044 (1/5) & +0.0169 (3/5) & +0.0081 (5/5) & +0.0117 & 5/6 \\
5 & Pooled descriptor & +0.0079 (4/5) & +0.0011 (3/5) & +0.0071 (3/5) & +0.0041 (4/5) & +0.0090 (3/5) & +0.0050 (3/5) & +0.0057 & 6/6 \\
6 & Auxiliary supervision, node & +0.0019 (2/5) & +0.0008 (3/5) & $-$0.0049 (3/5) & +0.0010 (3/5) & +0.0066 (3/5) & +0.0163 (5/5) & +0.0036 & 5/6 \\
6 & Auxiliary supervision, node + edge & +0.0086 (3/5) & $-$0.0023 (3/5) & +0.0007 (3/5) & $-$0.0035 (2/5) & $-$0.0111 (2/5) & +0.0246 (5/5) & +0.0028 & 3/6 \\
\midrule
--- & \textbf{EASE, mean} & \textbf{+0.0339 (5/5)} & \textbf{+0.0116 (5/5)} & \textbf{+0.0321 (5/5)} & \textbf{+0.0196 (5/5)} & +0.0404 (3/5) & \textbf{+0.0267 (5/5)} & \textbf{+0.0274} & 6/6 \\
\bottomrule
\end{tabular}}
\end{table*}

\begin{figure*}[t]
\centering
\includegraphics[width=\linewidth]{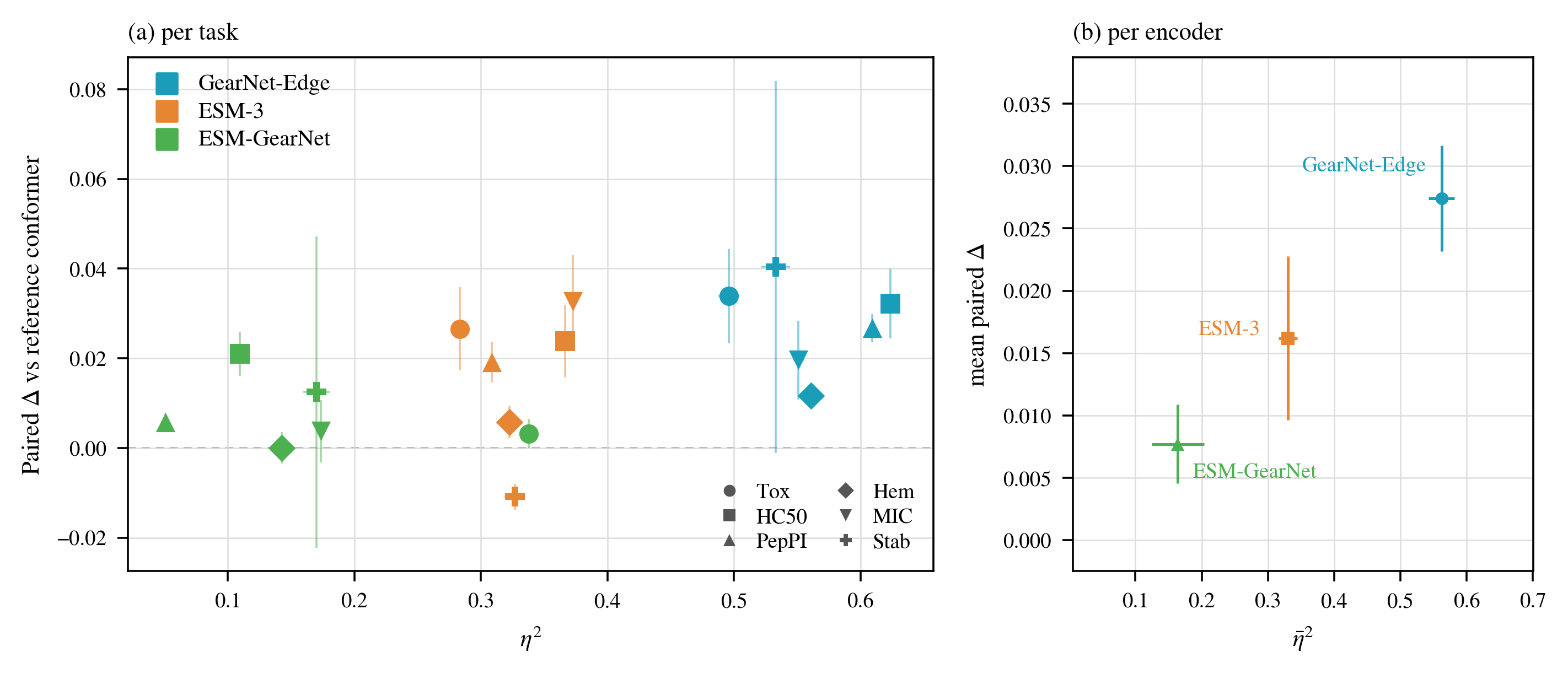}
\caption{\small
Relationship between conformer embedding variation and ensemble gains.
\textbf{(a)} Task-level comparison across three pretrained encoders and six peptide prediction tasks. The vertical axis shows the paired performance improvement of
EASE (mean fusion, $K=10$) over the reference conformer baseline.
\textbf{(b)} Encoder-level averages across the six tasks.
Error bars indicate $\pm 1$ standard error (bootstrap over peptides for x-axis in (a), across folds for y-axis in (a), and across tasks in (b)).}
\label{fig:eta-vs-gain}
\end{figure*}

Figure~\ref{fig:eta-vs-gain} shows how conformer embedding variations in different encoders influence EASE improvements over the reference conformer. Tables~\ref{tab:esmgearnet-abs} and ~\ref{tab:esmgearnet-paired} present the absolute and paired performance of EASE (mean $K=10$) and reference conformer on ESM-GearNet, compared to ESM-3 and GearNet-Edge.

\begin{table*}[t]
\centering
\small
\caption{Absolute test performance. Mean $\pm$ standard deviation over same folds. AUROC for classification, RMSE for regression. ESM-GearNet uses a frozen ESM-2-650M with LoRA on the GearNet trunk.}
\label{tab:esmgearnet-abs}
\resizebox{\textwidth}{!}{%
\begin{tabular}{llcccccc}
\toprule
Encoder & Arm & Toxicity $\uparrow$ & Hemolytic $\uparrow$ & HC50 $\downarrow$ & \textit{E. coli} MIC $\downarrow$ & Stability $\uparrow$ & PepPI $\uparrow$ \\
\midrule
ESM-GearNet & Reference conformer ($K{=}1$) & 0.7832 $\pm$ 0.0700 & 0.8895 $\pm$ 0.0252 & 0.5635 $\pm$ 0.0358 & 0.7059 $\pm$ 0.0168 & 0.6895 $\pm$ 0.0447 & 0.8352 $\pm$ 0.0097 \\
 & EASE, mean $K{=}10$ & 0.7864 $\pm$ 0.0722 & 0.8896 $\pm$ 0.0285 & 0.5425 $\pm$ 0.0418 & 0.7022 $\pm$ 0.0157 & 0.7020 $\pm$ 0.0474 & 0.8410 $\pm$ 0.0057 \\
\midrule
GearNet-Edge & Reference conformer ($K{=}1$) & 0.6893 $\pm$ 0.1008 & 0.8617 $\pm$ 0.0221 & 0.6460 $\pm$ 0.0191 & 0.6847 $\pm$ 0.0202 & 0.6737 $\pm$ 0.0458 & 0.8059 $\pm$ 0.0159 \\
 & EASE, mean $K{=}10$ & 0.7231 $\pm$ 0.0938 & 0.8733 $\pm$ 0.0225 & 0.6139 $\pm$ 0.0164 & 0.6651 $\pm$ 0.0275 & 0.7141 $\pm$ 0.0525 & 0.8327 $\pm$ 0.0101 \\
\midrule
ESM-3 & Reference conformer ($K{=}1$) & 0.7792 $\pm$ 0.0534 & 0.9013 $\pm$ 0.0209 & 0.5793 $\pm$ 0.0230 & 0.6842 $\pm$ 0.0225 & 0.7141 $\pm$ 0.0266 & 0.8329 $\pm$ 0.0118 \\
 & EASE, mean $K{=}10$ & 0.8058 $\pm$ 0.0387 & 0.9071 $\pm$ 0.0161 & 0.5554 $\pm$ 0.0334 & 0.6515 $\pm$ 0.0183 & 0.7032 $\pm$ 0.0314 & 0.8520 $\pm$ 0.0100 \\
\bottomrule
\end{tabular}%
}
\end{table*}

\begin{table*}[t]
\centering
\small
\caption{Paired difference of EASE with mean $K{=}10$ against the reference conformer over the same folds (folds improved). Training seeds are averaged within a fold before pairing, and the sign is set so positive always indicates improvement for both AUROC and RMSE. The Mean column averages the six tasks.}
\label{tab:esmgearnet-paired}
\resizebox{\textwidth}{!}{%
\begin{tabular}{llccccccc}
\toprule
Encoder & Toxicity & Hemolytic & HC50 & \textit{E. coli} MIC & Stability & PepPI & Mean \\
\midrule
ESM-GearNet & $+0.0032$ (3/5) & $+0.0000$ (3/5) & $+0.0210$ (5/5) & $+0.0037$ (3/5) & $+0.0125$ (2/5) & $+0.0058$ (4/5) & $+0.0077$ \\
\midrule
GearNet-Edge & $+0.0339$ (5/5) & $+0.0116$ (5/5) & $+0.0321$ (5/5) & $+0.0196$ (5/5) & $+0.0404$ (3/5) & $+0.0267$ (5/5) & $+0.0274$ \\
\midrule
ESM-3 & & $+0.0266$ (5/5) & $+0.0058$ (4/5) & $+0.0238$ (4/5) & $+0.0327$ (5/5) & $-0.0108$ (0/5) & $+0.0191$ (5/5) & $+0.0162$ \\
\bottomrule
\end{tabular}%
}
\end{table*}

\end{document}